\documentclass[11pt]{article}

\usepackage[preprint]{acl}

\usepackage{times}
\usepackage{latexsym}
\usepackage[T1]{fontenc}
\usepackage[utf8]{inputenc}
\usepackage{microtype}
\usepackage{inconsolata}
\usepackage{graphicx}
\usepackage{hyperref}
\usepackage{url}
\usepackage{booktabs}
\usepackage{enumitem}
\usepackage{wrapfig}
\usepackage{algorithm}
\usepackage{algpseudocode}
\usepackage[misc]{ifsym}
\usepackage{subcaption}
\usepackage{microtype}
\usepackage{amsmath}
\usepackage{colortbl}
\usepackage{xcolor} 
\definecolor{lightgray}{rgb}{0.9,0.9,0.9}
\usepackage{caption}
\usepackage{setspace}
\usepackage{url}
\usepackage{multirow}
\usepackage{tabularx}
\usepackage{blindtext}
\usepackage{pgfplots}
\pgfplotsset{compat=1.18} 
\usepackage{tikz}
\usetikzlibrary{er,positioning,bayesnet}
\usepackage{makecell}
\usepackage{tipa}
\usepackage{siunitx}
\usepackage{nicefrac}
\usepackage{listings}
\usepackage[raster,skins, most]{tcolorbox} %
\usepackage{xltabular}
\usepackage{adjustbox}
\usepackage{xurl}
\usepackage{rotating}
\usepackage[normalem]{ulem}
\usepackage{epstopdf}
\usepackage{multirow}
\usepackage{tabularx}
\usepackage{pifont}
\usepackage{amssymb}
\usepackage{fvextra}
\DefineVerbatimEnvironment{verbatim}{Verbatim}{
  breaklines=true,
  breakanywhere=true,
  breaksymbolleft={}
}

\usepackage[table]{xcolor}
\definecolor{degradePink}{HTML}{F8D7DA}   
\definecolor{maxGainPurple}{HTML}{E7DAFF} 

\usepackage[table]{xcolor}
\definecolor{plotblue}{HTML}{6492F0}

\newtcolorbox{mybox}[2][]
  {colback = black!5!white, colframe = black!75!black, fonttitle = \bfseries,
    colbacktitle = black!100!black, enhanced, before upper={\fontsize{8}{11}\obeyspaces\obeylines\selectfont}, fontupper=\selectfont,
    attach boxed title to top left={yshift=-2.2mm,xshift=4mm},
    title=#2,#1}
\definecolor{purple1}{RGB}{126, 107, 196}

\newcommand{\ours}{DeepSearch-World}
\newcommand{\framework}{DeepSearch-Evolve}

\title{DeepSearch-World: Self-Distillation for Deep Search Agents in a Verifiable Environment}

\author{
Xinyu Geng\textsuperscript{1,2,*},
Xuanhua He\textsuperscript{1,*},
Sixiang Chen\textsuperscript{2,3,*},
Yanjing Xiao\textsuperscript{1},
Fan Zhang\textsuperscript{2},
\\
\textbf{
Shijue Huang\textsuperscript{1},
Haitao Mi\textsuperscript{2},
Zhenwen Liang\textsuperscript{2,\textdagger},
Tianqing Fang\textsuperscript{2,\textdagger},
Yi R. Fung\textsuperscript{1,\textdagger}}
\vspace{5pt}
\\
\textsuperscript{1}HKUST, \quad
\textsuperscript{2}Tencent, \quad
\textsuperscript{3}HKUST(GZ)
\\
\small{
\href{https://huggingface.co/Ornamentt/deepsearch-code/tree/main}{Code}
\quad|\quad
\href{https://huggingface.co/datasets/Ornamentt/deepsearch-world-data}{Data}
\quad|\quad
\href{https://huggingface.co/datasets/Ornamentt/DeepSearch-World-Env}{Env}
}
\\
\small{
\textsuperscript{*}Equal contribution
\quad
\textsuperscript{\textdagger}Corresponding authors
}
}

\begin{document}
\maketitle

\begin{abstract}


Training tool-use agents to improve from their own experience remains challenging, as supervised fine-tuning relies on fixed teacher-distilled trajectories, while sparse-reward reinforcement learning provides weak supervision for long-horizon interactions. 
We present \framework{}, a self-distillation framework for web agents built on \ours{}, a deterministic and verifiable environment with \textit{reproducible} search and page-reading tools. 
\ours{} contains 420K multi-hop QA tasks constructed from entity-level random walks and supports key agentic cognitive behaviors useful for self-evolving, including progress verification, grounded reflection, and failure recovery. 
\framework{} iteratively performs trajectory generation, filtering, data mixing, and fine-tuning to train stronger agents. 
Without distillation from more capable models, \ours{}-9B achieves competitive performance compared with open-source agents, 
reaching 31.2\% on BrowseComp, 61.5\% on GAIA, and 93.4\% on HotpotQA, 
showing that verifiable environments enable scalable self-evolution for long-horizon web agents. 
We will release the environment, 420K training pool, validation set, model, and code to facilitate future research on self-improving deep search agents.
\end{abstract}

\section{Introduction}



Recent advances in large language models (LLMs) have enabled agents that move beyond passive text generation to plan multi-step tasks, formulate search queries, read documents, browse the web, and refine answers through tool use and iterative reasoning~\citep{schick2024toolformer,deepresearch,geminideepresearch}. However, enabling these agents to improve autonomously from their own interactions remains a key challenge toward scalable self-evolving agents~\citep{he-etal-2025-openwebvoyager}.

A common recipe is supervised fine-tuning (SFT) on positive trajectories distilled from the backbone model itself~\citep{zelikman2022star,zeng2024agenttuning}. 
However, the performance easily saturates after a few optimization steps, bounded by the inherent capability of the backbone model itself and the diversity of the self-collected trajectories~\citep{song2025mind}.
Another way to instantiate self-evolving is to optimize tool-use agents on their own rollouts with verifiable rewards, typically using RL-style objectives such as GRPO~\citep{guo2025deepseek,team2025kimi,team2025tongyi}. However, these rewards are typically sparse and trajectory-level, offering little guidance on whether failures arise from query formulation, tool selection, evidence extraction, or answer synthesis~\citep{liu2025agentic,li2026salt}.

On-policy self-distillation (OPSD) alleviates reward sparsity by matching the student's own rollouts to dense token-level distributions from a fine-grained teacher policy~\citep{hubotter2026reinforcement,ye2026policy,shen2026geometry}. 
However, in agentic tool use, such fine-grained supervision lies in actions, including tool selection, evidence verification, search query reformulation, and progress tracking~\citep{gritta2026process,liu2025veriweb}, which 
requires a deterministic and verifiable interaction environment. Otherwise, the teacher distribution at each tool-use step can be noisy, limiting the direct applicability of OPSD to long-horizon agents.

\begin{figure*}[t]
    \centering
    \includegraphics[width=0.95\textwidth]{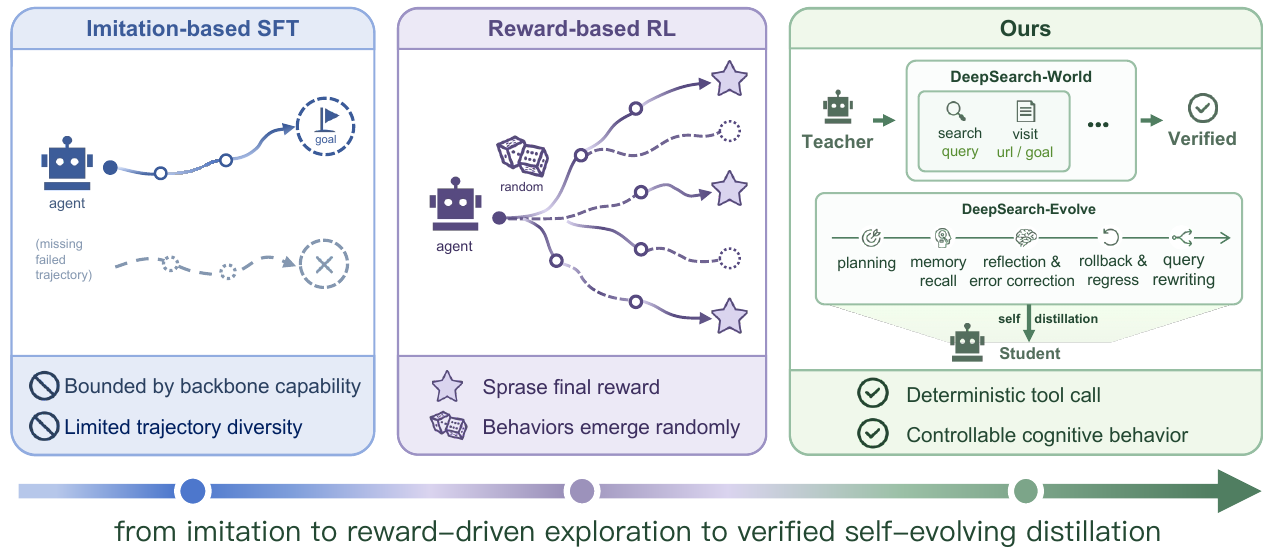}
    \caption{\textbf{Conceptual comparison of self-evolving paradigms for tool-use agents.} SFT imitates positive traces, reward-based RL learns from sparse outcomes, whereas our \framework{} distills verified tool-use behaviors in a deterministic environment \ours{}, providing controllable process-level supervision for planning, memory, error correction, rollback, and query rewriting.}
    \label{fig:intro}
\end{figure*}

In summary, as shown in Fig.~\ref{fig:intro}, existing self-evolving pipelines for tool-use agents are limited by static imitation signals, sparse outcome rewards, or unreliable dense supervision. 
\textbf{Long-horizon agents instead require a verifiable environment that can expose process-level supervision over intermediate tool-use decisions.}
Therefore, we introduce \ours{}, a deterministic and verifiable environment for deep search agents. 
Built on \ours{}, we further develop \framework{}, a self-distillation framework that enables agents to iteratively improve from their own verified tool-use experience.
This work makes three contributions:

First, we introduce \ours{}, a deterministic and verifiable offline environment for deep-search agents with search and browse tools, accompanied by 420K aligned multi-hop QA tasks over Wikipedia. Unlike live web environments, \ours{} provides reproducible observations and entity-level verification for intermediate tool-use progress. Second, we propose scaffold process supervision for long-horizon tool use. The teacher explicitly tracks progress, evidence, failed attempts, and recovery during rollout, and we convert these scaffold trajectories into standard ReAct-format supervision for student learning, injecting planning, memory tracking, grounded reflection, and failure recovery into student. Third, we develop an evolving-SFT framework where agents iteratively generates trajectories, receive verified process signals, and iteratively absorb successful behaviors. This enables open-source agents to improve from verified experience without relying on synthetic trajectories from stronger proprietary models.

Experimental results show that \ours{}-9B achieves competitive performance among open-source agents without distillation of strong proprietary models.
Additional analyses show that \ours{}-9B sustains longer tool-use interactions, uses tools more effectively, and benefits from both verified trajectory filtering and scaffold-to-ReAct conversion.

\begin{figure*}[t]
\centering
  \centering
  \includegraphics[width=\linewidth]{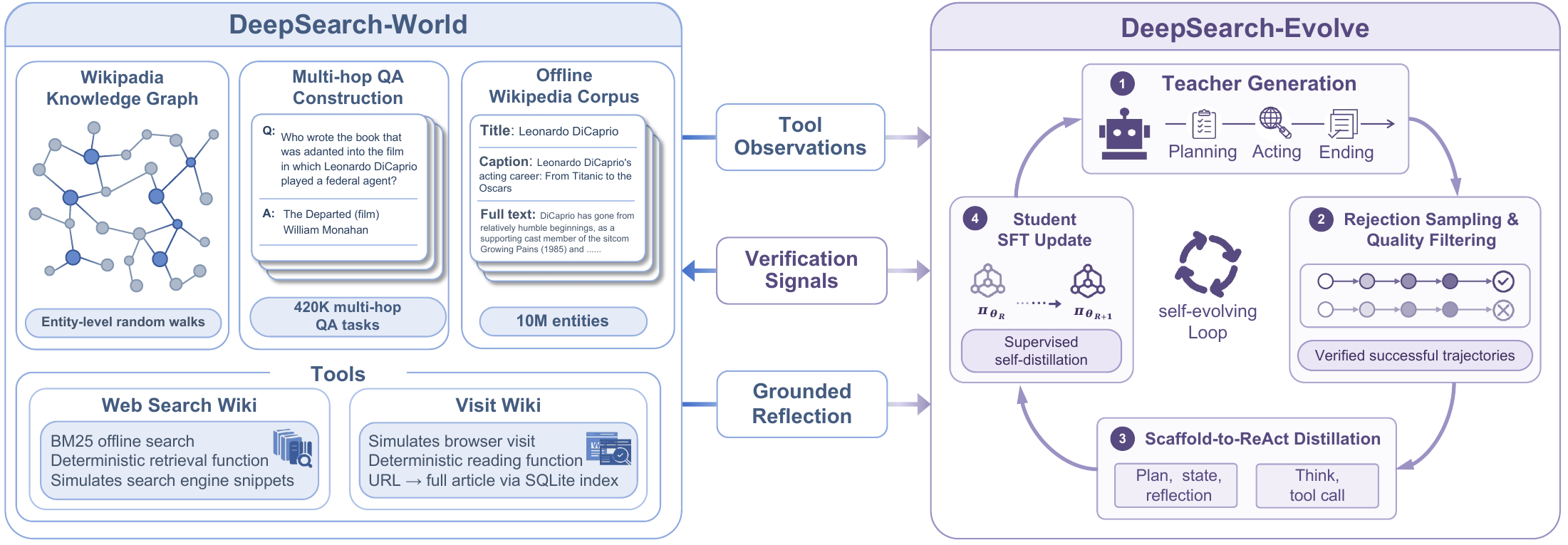}
  \caption{\textbf{Overview of the \ours{} and \framework{}.} \ours{} provides a deterministic and verifiable tool environment, while \framework{} performs self-distillation through a self-evolving loop of trajectory generation, rejection sampling, importance-sampling-based data mixing, and SFT training.}
  \label{fig:overview}
\end{figure*}

\section{Related Work}
\label{sec:related_work}

\paragraph{Tool-Use Agents and Environments}
Tool-augmented LLM agents extend model capabilities through external search, API invocation, and environment interaction~\citep{yao2023react,schick2024toolformer,chen2026efficient}. Prior work explores large-scale tool-use training, web navigation agents, and trajectory-level agent tuning~\citep{li2025websailor,geng2026geobrowse,chen2026genevolve,su2026agentvista}. 
To improve reproducibility and scalability, recent studies investigate virtual tool environments through API caching, programmatic simulators, or LLM-based environment modeling~\citep{guo2024stabletoolbench,guo2025mirrorapi,li2025simulating}. However, these approaches suffer from limited realism, hallucinated observations, or high serving cost. In contrast, our approach constructs a deterministic offline environment grounded in Wikipedia, enabling scalable and reproducible agent trajectory generation.

\paragraph{Self-evolving Agents}
Self-improvement methods iteratively refine model behavior using model-generated trajectories and verification signals~\citep{zelikman2022star,singh2024humanlevel,yuan2023scaling}. 
GRPO-style group RL has become the dominant post-training paradigm for tool-use agents~\citep{team2025tongyi,geng2025webwatcher,huang2026towards}, but its reliance on sparse rewards and weak verifiability makes optimization unstable. Recent OPSD further seek to alleviate this distilling from on-policy generation~\citep{hubotter2026reinforcement,zhao2026self,ye2026policy}, but the teacher policy does not always provide a reliable or stationary target distribution. We address this by introducing an evolving SFT method and a verifiable tool environment that enables stable self-distillation.
\section{Methodology}
\label{sec:methodology}

As shown in Fig.~\ref{fig:overview}, we present \textbf{\framework{}}, a self-distillation framework for search agents consisting of three key components: (1)~a verifiable tool environment \textbf{\ours{}} and multi-hop QA data construction, (2)~a scaffold teacher agent that generates high-quality trajectories, and (3)~an iterative self-evolving training loop with asynchronous generation and training.

\subsection{Verifiable Tool Environment}
\label{sec:environment}

\subsubsection{Data and Environment Construction}
\label{subsec:construction}
To support scalable self-distillation, we build a deterministic offline Wikipedia environment \textbf{\ours{}} together with a large-scale multi-hop QA dataset. Let $\mathcal{G}=(\mathcal{V},\mathcal{E})$ be the Wikipedia hyperlink graph. Following~\citet{geng2025webwatcher}, we sample entity-level random walks $\tau=(v_1,\ldots,v_H)$ from $\mathcal{G}$, where each node corresponds to a target entity and the walk defines an $H$-hop reasoning chain. We construct 420K QA instances by obfuscating explicit entity mentions, requiring the agent to recover the target entities through reasoning, search-query revision, and evidence discovery. A large task pool broadens the induced trajectory distribution, covering diverse reasoning paths, failure modes, and recovery patterns. We analyze this effect in Sec.~\ref{sec:datascale}.

We then ground the QA tasks in an offline corpus to instantiate \ours{}. For each instance $q_i$, let $\mathcal{T}_i=\{e_{i,1},\ldots,e_{i,H}\}$ be the target entity set defined by its random walk. We crawl the corresponding Wikipedia pages for $\bigcup_i \mathcal{T}_i$ and build a local corpus $\mathcal{C}$ of approximately 10 million entries, each consisting of a title, a caption, and full article text. This alignment ensures that the evidence required by each question is searchable and verifiable within the offline environment. Implementation details are provided in Appendix~\ref{app:environment_details}.

From the 420K QA pool, we reserve a held-out validation split, \textbf{DeepSearch-Val}, for training validation and behavioral analysis. 
It contains 377 high-quality instances whose required evidence is covered by the offline corpus and retrievable through the provided search and visit tools, enabling reliable evaluation in the deterministic environment \ours{}. 
These instances are excluded from trajectory generation and model training. 
Answers are verified by a five-expert pool, with each instance independently validated by at least three experts.


\subsubsection{Offline tools}


The environment exposes two offline tools that follow the standard search--read workflow of web agents while keeping all observations local, deterministic, and verifiable. Formally, let $\mathcal{I}$ denote the Lucene BM25 index built over Wikipedia corpus $\mathcal{C}$ with Pyserini~\citep{lin2021pyserini}. The search tool is a deterministic retrieval function $\texttt{web\_search\_wiki}(x)\mapsto \{(t_j,s_j,u_j)\}_{j=1}^{K}$，
where $x$ is a natural-language query, and each returned candidate consists of a page title $t_j$, a short caption $s_j$, and a deterministic URL $u_j$. This simulates a search engine that returns snippets and links rather than full documents. The visit tool is a deterministic reading function
$\texttt{visit\_wiki}(u_j)\mapsto d_j$
which maps a URL to the corresponding full article $d_j\in\mathcal{C}$ through a SQLite offset index, simulating a browser visit for reading detailed evidence.

Although the backend is restricted to Wikipedia, the tool schema is aligned with real web tools: search maps queries to ranked snippets and URLs, while visit maps URLs to page content. This schema alignment allows the offline tools to be replaced by live web search and visit tools without changing the action format. Implementation details are provided in Appendix~\ref{app:virtual_tool_details}.

\subsubsection{Environment-grounded reflection}


For each question $q_i$, \ours{} stores a ground-truth entity set $\mathcal{T}_i=\{e_{i,1},\ldots,e_{i,H}\}$ for process-level verification. During rollout, the environment maintains a completed set $\mathcal{S}_t\subseteq\mathcal{T}_i$. A tool response $o_t$ is considered successful if it matches any unresolved entity in $\mathcal{T}_i\setminus\mathcal{S}_t$, after which the matched entity is added to $\mathcal{S}_{t+1}$.
This order-free verification identifies objective progress after each tool call without expensive LLM judgments.
Failed calls trigger staged rule-based reflection toward the next unresolved entity. The first failure receives generic revision signals, while repeated failures reveal stronger guidance such as the canonical entity name or a fuzzy description. This yields grounded ``search--fail--reflect--retry'' trajectories and encourages query reformulation and recovery. Reflection is used only during scaffold teacher rollout and is removed or rewritten as self-correction for student.

\subsection{Scaffold Teacher Agent}
\label{sec:agent}

Following~\citet{fang2026cognitivekernelproframeworkdeep}, we create a scaffold teacher with three phases: \textsc{Plan}, \textsc{Act}, and \textsc{End}. The scaffold is only for trajectory generation, explicitly supervising planning, memory, error recovery, and grounded answering before distillation.

\begin{figure}[t]
    \centering
    \includegraphics[width=\columnwidth]{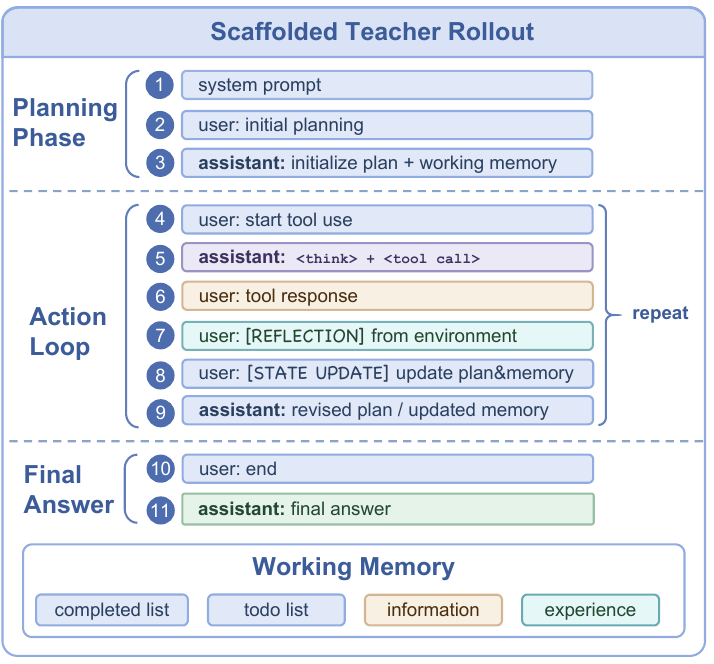}
    \caption{\textbf{Three-phase scaffolded teacher rollout.} 
The teacher proceeds through the initial \textit{Plan}, iterative \textit{Act}, and \textit{End} period.}
    \label{fig:scaf}
\end{figure}

\paragraph{Plan.}
Given a question $q$, the teacher initializes a structured progress state $s_0$ for task decomposition and evidence tracking. 
Fig.~\ref{fig:scaf} shows the state contains four writable fields: \texttt{completed\_list} for confirmed subgoals, \texttt{todo\_list} for following actions, \texttt{experience} for useful lessons from failed attempts, and \texttt{information} for evidence extracted from observations.

\begin{figure*}[t]
    \centering
    \includegraphics[width=\textwidth]{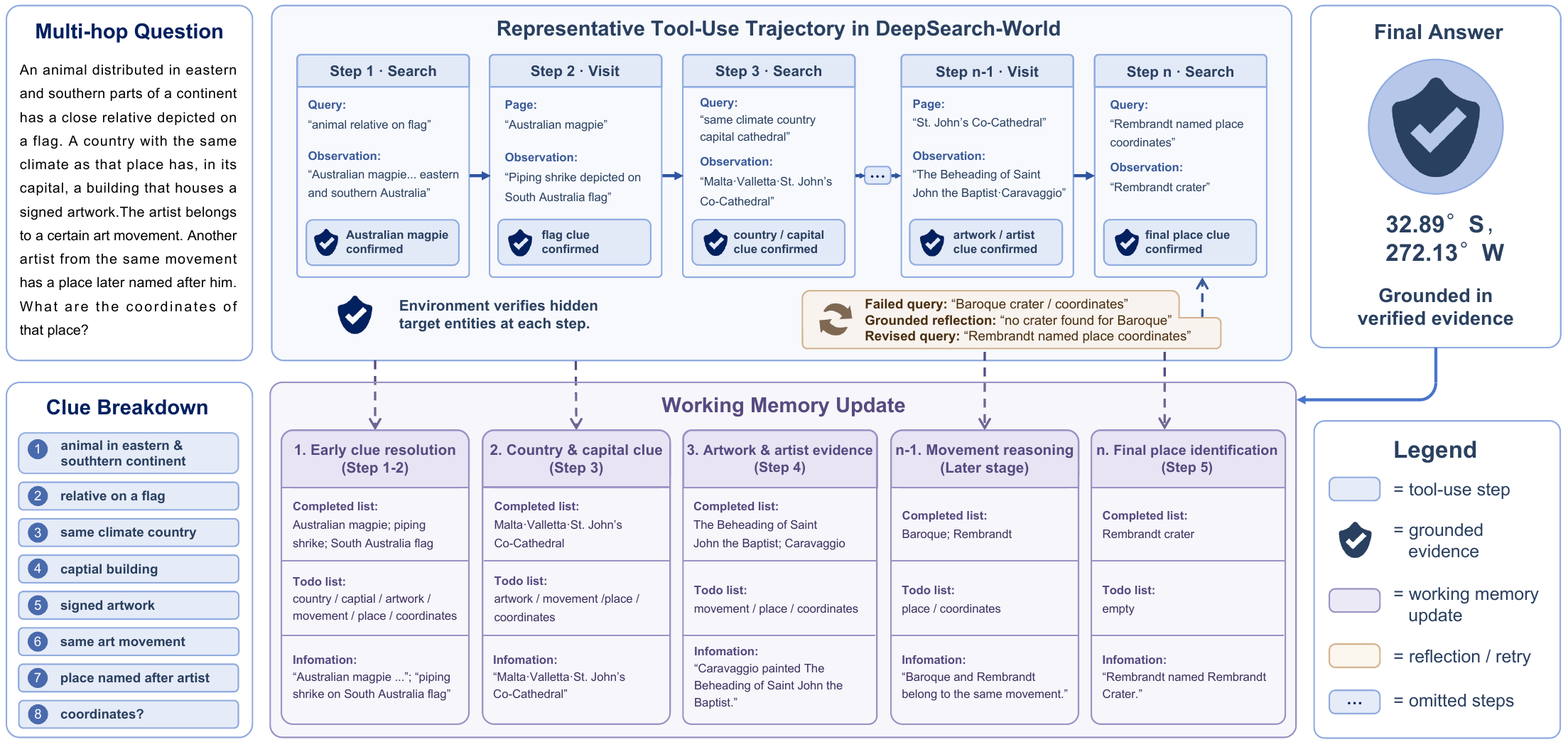}
    \caption{\textbf{Example scaffold trajectory generated by the teacher agent.} 
The teacher decomposes a fuzzy multi-hop question, alternates between search and visit tools, updates working memory with verified evidence, uses grounded reflection to recover from failed searches, and produces a final answer supported by verified observations.}
    \label{fig:case1}
\end{figure*}

\paragraph{Act.}
The teacher then performs up to $T_{\max}$ interaction steps. 
At step $t$, it selects a tool call $a_t$ from the current progress state $s_t$, receives an observation $o_t$, and updates
\begin{equation}
    s_{t+1}=\mathcal{U}(s_t,a_t,o_t,r_t),
\end{equation}
where $r_t$ is the environment-grounded reflection returned upon retrieval failure. 
This process records verified evidence, failed strategies, and revised goals, yielding trajectories with both successful tool use and recoverable failures.

\paragraph{End.}
When evidence is sufficient or the budget is exhausted, the teacher enters \textsc{End} and generates a concise answer grounded in verified working memory, reducing unsupported synthesis and hallucination. An example rollout is shown in Fig.~\ref{fig:case1}.

\subsection{Self-Evolving Training Loop}
\label{sec:training_loop}


\framework{} uses an iterative self-evolving loop that alternates between scaffold trajectory generation and training. 
At round $R$, the current model $\pi_{\theta_R}$ generates trajectories in the verifiable virtual-tool environment as the teacher. Then verified successful trajectories are converted into ReAct-format supervision to train the student $\pi_{\theta_{R+1}}$, which will be the next teacher. 
This enables the agent to improve from its own verified tool-use experience rather than relying on a fixed demonstration set.

\paragraph{Trajectory generation and verification.}
For each question $q$, $\pi_{\theta_R}$ generates a scaffold trajectory $\tau$ with the teacher agent in Sec.~\ref{sec:agent}, including tool calls, observations, state updates, reflections, and a final answer. 
We retain trajectories with correct answer and further prompt Qwen3.5-9B~\citep{qwen3.5} to apply trajectory-level filtering to remove redundant evidence, weak goal alignment, and inconsistent reasoning. The prompt we use is in Appendix~\ref{app:prompt_filter}.

\paragraph{Scaffold-to-ReAct conversion.}
The scaffold improves teacher trajectory quality but is not exposed to the student, since our goal is to train a deployable ReAct agent rather than a model dependent on external planning or reflection prompts. 
Each retained scaffold trajectory is therefore converted into standard ReAct-format supervision by removing stage-specific prompts and rewriting progress states and reflections into the assistant reasoning trace.

For each tool-use step $t$, the target \texttt{<think>} block is constructed as
\begin{equation}
    \texttt{<think>}_t
    =
    P_t \oplus R_t \oplus A_t .
\end{equation}
Here, $P_t$ is rewritten from the current progress state $s_t$, summarizing completed subgoals, pending targets, failed-search experience, and verified evidence; $R_t$ rewrites the environment reflection $r_t$ as self-correction for query adjustment; and $A_t$ preserves the local rationale for action $a_t$. 
The action $a_t$ and observation $o_t$ are kept unchanged. 
This conversion trains a standard ReAct policy while distilling scaffold-induced planning, memory tracking, and error recovery. 
An example converted trajectory is provided in Appendix~\ref{app:case}.

\begin{table*}[t]
\centering
\scriptsize
\setlength{\tabcolsep}{4.5pt}
\resizebox{\textwidth}{!}{
\begin{tabular}{l|cccccc}
\toprule
\textbf{Model / Agent}
& \textbf{BrowseComp}
& \textbf{BrowseComp-ZH}
& \textbf{HLE}
& \textbf{GAIA}
& \textbf{xbench}
& \textbf{HotpotQA} \\
\midrule

\rowcolor{gray!33}
\multicolumn{7}{c}{\emph{\textbf{Proprietary Agents}}} \\
\midrule
OpenAI Deep Research    & \textbf{51.5} & 42.9 & \textbf{26.6} & 67.4 & --   & -- \\
OpenAI-o3               & 49.7 & \textbf{58.1} & 20.2 & \textbf{70.5} & \textbf{65.0} & -- \\

\midrule
\rowcolor{gray!33}
\multicolumn{7}{c}{\emph{\textbf{Open Source Agents}}} \\
\midrule
R1-Searcher          & 1.0 & -- & 5.4  & 8.3   & --   & 62.4 \\
Search-R1            & 0.4 & -- & 13.0 & 18.7  & --   & 63.2 \\
ZeroSearch           & 1.4 & -- & 8.6  & 9.9   & --   & 32.4 \\
ASearcher            & 3.2 & -- & 13.8 & 22.1  & --   & 71.0 \\
DeepResearcher       & 1.8 & -- & 6.0  & 24.0 & --   & 56.6 \\
PokeeResearch        & 6.2 & -- & 17.6 & 49.2  & --   & 71.6 \\
WebSailor            & 6.7  & 14.2 & 12.8 & 37.9 & 34.3 & 92.8 \\
OffSeeker-DPO        & 12.8 & 26.6 & 17.6 & 51.5  & 48.0 & -- \\
WebExplorer          & 15.7 & 32.0 & 17.3 & 50.0 & 53.7 & -- \\
Marco-DR             & 31.4 & 47.1 & --   & 69.9 & 42.0 & -- \\
MiroThinker-v1.0     & 31.1 & 40.2 & 21.5 & 66.4 & 34.0 & -- \\
DeepDive             & 6.3  & 15.1 & --   & --   & 35.0 & -- \\
\midrule
\rowcolor{gray!33}
\multicolumn{7}{c}{\emph{\textbf{Ours}}} \\
\midrule
Qwen3.5-9B-Instruct     & 7.4 & 13.5 & 16.7 & 23.9 & 20.0 & 45.3 \\
\rowcolor{maxGainPurple!40}
\textsc{\ours{}-9B}          & 31.2 & 36.4 & 25.7 & 61.5 & 49.0 & \textbf{93.4} \\
\rowcolor{maxGainPurple!40}
$\Delta$                & \textcolor{blue}{+23.8}  & \textcolor{blue}{+22.9}  & \textcolor{blue}{+9.0}   & \textcolor{blue}{+37.6} & \textcolor{blue}{+29.0} & \textcolor{blue}{+48.1} \\

\bottomrule
\end{tabular}
}
\caption{Main results on deep search and related reasoning benchmarks. All scores are reported as percentages. A dash indicates that the corresponding result is unavailable or not applicable.}
\label{tab:main_results}
\end{table*}

\paragraph{Supervised self-distillation update.}
We update the agent with SFT on environment-verified ReAct trajectories, which can be viewed as hard-label token-level policy distillation. 
Given a verified trajectory $\tilde{\tau}=(x,y_{1:T})$, SFT minimizes
\begin{equation}
\begin{aligned}
\mathcal{L}_{\mathrm{SFT}}(\theta)
=
\mathbb{E}_{\tilde{\tau}\sim \widetilde{\mathcal{D}}^{(R)}}
\sum_{t=1}^{T}
\mathrm{KL}\big(
\delta_{y_t}
\,\Vert\,
\pi_\theta(\cdot \mid x,y_{<t})
\big),
\end{aligned}
\end{equation}
which is equivalent to negative log-likelihood up to a constant. 
Meanwhile, on-policy self-distillation (OPSD) matches the student to a soft teacher distribution on student-induced prefixes. Let
$\pi_\theta^t=\pi_\theta(\cdot \mid x,\hat{y}_{<t})$
and
$q^t=q(\cdot \mid x,\hat{y}_{<t})$.
Then OPSD minimizes
\begin{equation}
\mathcal{L}_{\mathrm{OPSD}}(\theta)
=
\mathbb{E}_{x,\,\hat{y}_{<t}\sim \pi_\theta}
\sum_t
\mathrm{KL}\big(q^t \,\Vert\, \pi_\theta^t\big).
\end{equation}
Thus, both objectives provide dense token-level supervision, but differ in prefix and target distributions. SFT uses verified offline prefixes with hard targets, whereas OPSD uses on-policy prefixes with soft targets.

We adopt evolving SFT for stability in long-horizon tool use. 
Fully on-policy distributional supervision is costly and can be unreliable, as student rollouts may drift into low-quality tool states where teacher token distributions become noisy or misleading. 
Instead, we sample scaffold rollouts from the current agent, filter them through rejection sampling and filtering, and convert them into ReAct traces. 
These traces supervise planning, tool invocation, error recovery, and final synthesis without relying on per-token guidance over unverified states. 
The resulting update is partially on-policy in data generation but teacher-forced in optimization, trading full distributional matching for stable, efficient, and reusable supervision.

\begin{figure*}[t]
    \centering
    \includegraphics[width=\textwidth]{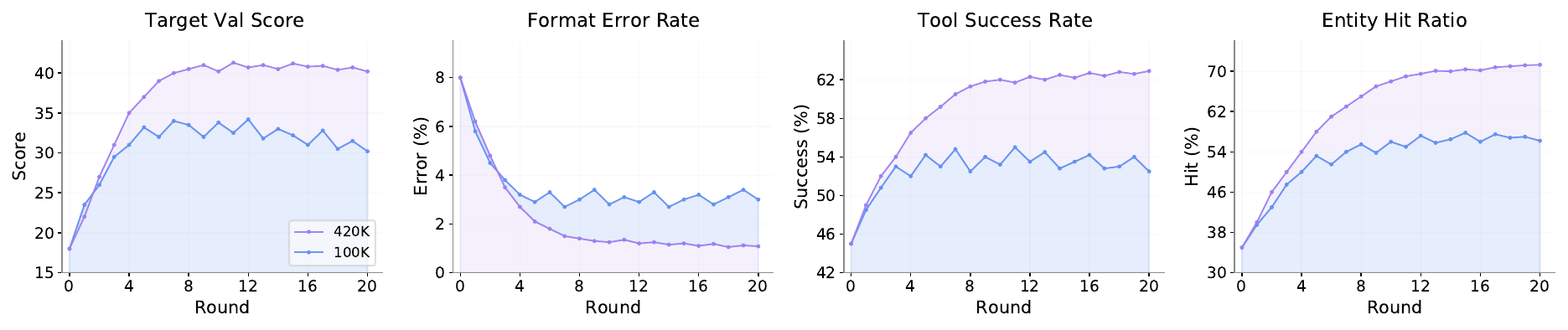}
    \caption{Data scale effect in evolving SFT.
Compared with 100K total training data, 420K QA lead to a higher target-validation plateau, lower residual format errors, and more stable gains in tool success and entity hit ratio.}
    \label{fig:scaffold}
\end{figure*}

\section{Experiments}
\label{sec:experiments}

\subsection{Experimental Setup}

\paragraph{Datasets.} We evaluate on seven deep search and related reasoning benchmarks: BrowseComp~\citep{wei2025browsecompsimplechallengingbenchmark}, BrowseComp-ZH~\citep{zhou2025browsecomp}, HLE~\citep{phan2025lastexam}, GAIA~\citep{mialon2023gaia},
  xbench~\citep{chen2025xbench},
  HotPotQA~\citep{yang2018hotpotqa} and Search-QA~\citep{jin2025search}. 
  Details of these benchmarks are provided in
  Appendix~\ref{app:benchmark_details}.

\paragraph{Baselines.} We compare against proprietary and open-source models under two evaluation settings. \emph{Proprietary Agents} includes OpenAI Deep Research~\citep{deepresearch} and OpenAI-o3~\citep{o3}.
\emph{Open Source Agents} test PokeeResearch \citep{wan2026rethinkingdesignreinforcementlearningbased}, WebSailor~\citep{li2025websailor}, Marco-DR~\citep{zhu2026marco}, MiroThinker-v1.0~\citep{team2025mirothinker}, WebExplorer~\citep{liu2025webexplorer}, DeepDive~\citep{lu2025deepdive}, R1-Searcher~\citep{song2025r1searcher}, Search-R1~\citep{jin2025searchr1}, ZeroSearch~\citep{sun2025zerosearch}, ASearcher~\citep{gao2025asearcher}, Offseeker~\citep{zhou2026offseeker}, and DeepResearcher~\citep{zheng2025deepresearcher}.
Finally, we compare \ours{}-9B with Qwen3.5-9B-Instruct backbone.

\paragraph{Training Configuration.}
We train \ours-9B from Qwen3.5-9B for 11 self-evolving rounds in \ours. 
Each round generates trajectories for 10,000 instances and triggers training once 4,000 trajectories pass rejection sampling and quality filtering, with trajectory generation capped at $T_{\max}=30$ steps. 
Since generation and training are asynchronous, we use importance sampling when new trajectories outpace SFT updates, sampling across rounds with decay factor $\gamma=0.5$ to prioritize recent verified data while retaining earlier trajectories to mitigate catastrophic forgetting. 
We fine-tune for one epoch per round with Llama Factory~\citep{zheng2024llamafactory}. 
To reduce the offline-to-real execution gap, we further apply GRPO on 1,600 real-tool instances, using Google SerpAPI~\citep{serpapi} for search and Jina~\citep{jina} for page retrieval. 
Full hyperparameters are provided in Appendix~\ref{app:train_details}.

\subsection{Main Results}
\label{sec:main_results}

Tab.~\ref{tab:main_results} compares \ours{}-9B with proprietary systems and recent open-source deep-search agents. 
A key distinction is the supervision source. Many competitive open-source agents rely on stronger-model trajectories or multi-agent synthetic pipelines, e.g., O-Researcher uses multi-agent distillation, and Marco-DR synthesizes SFT trajectories from frontier foundation models.  In contrast, \ours{}-9B is optimized only from its own environment-verified rollouts and achieves competitive open-source performance. 
This suggests that deterministic environment verification can replace part of the supervision traditionally supplied by stronger teachers or external synthetic pipelines. 

Under the same tool setting, \ours{}-9B consistently improves over Qwen3.5-9B-Instruct across all benchmarks, with gains of +23.8 on BrowseComp, +22.9 on BrowseComp-ZH, +9.0 on HLE, +37.6 on GAIA, +29.0 on xbench, and +48.1 on HotpotQA. 
These gains indicate that evolving self-distillation learns transferable tool-use behaviors, including query reformulation, evidence grounding, and multi-step synthesis. The lower BrowseComp-ZH score is expected because training uses only English trajectories, though the improvement over the backbone still indicates partial cross-lingual transfer.

\subsection{Analysis}

\subsubsection{Self-Evolving Trajectory Quality Over Rounds}
\label{sec:datascale}

Fig.~\ref{fig:scaffold} compares evolving SFT under different data-pool scales, with each round trained on 4K verified trajectories. 
The 420K pool provides sufficient unseen instances across rounds, whereas the 100K pool is exhausted much earlier, causing later trajectory generation to rely on previously rejected or low-yield instances with weaker supervision. 
Consequently, the 420K setting reaches a higher validation plateau, lower residual format error, and more stable gains in tool success and entity hit ratio, while the 100K setting improves early but saturates sooner and fluctuates more.
This suggests that evolving SFT relies on data-pool diversity rather than repeated exposure. 
Duplicated trajectories risk overfitting to narrow interaction patterns, whereas a larger pool better covers world knowledge, task forms, and failure-recovery behaviors, improving generalization and preserving agentic skills across rounds.
Based on this trend, we use 11 evolving rounds for the main experiments, which provides strong validation performance before the curve largely saturates.

\subsubsection{Tool-Use Behavior Analysis}

\begin{figure*}[t]
\vspace{-10pt}
\centering

\begin{minipage}[t]{0.62\textwidth}
  \vspace{0pt}
  \centering
  \includegraphics[width=0.95\linewidth]{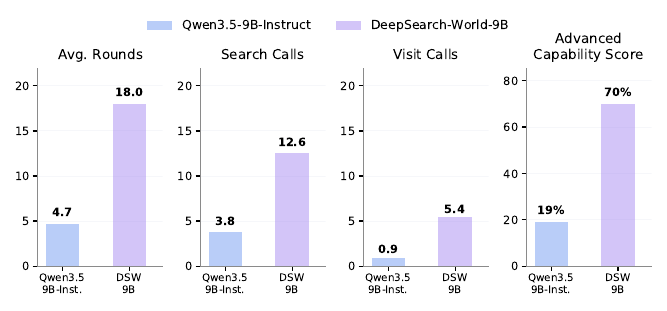}
  \vspace{-4pt}
  \captionof{figure}{Tool-use behavior and advanced capability on \textsc{DeepSearch-Val}. 
  Compared with Qwen3.5-9B-Instruct, \ours{} sustains longer interactions, performs more \texttt{search} and \texttt{visit} calls, and achieves a substantially higher advanced capability score.}
  \label{fig:tool}
\end{minipage}
\hfill
\begin{minipage}[t]{0.35\textwidth}
  \vspace{0pt}
  \centering
  \captionof{table}{Ablation of rejection sampling and trajectory quality filtering. 
  \textbf{RS only} selects answer-correct trajectories, \textbf{QF only} applies the trajectory quality filter, and \textbf{w/o both} removes both. 
  Results are in percentages averaged over three runs.}
  \label{tab:ablation_quality_filter}
  \vspace{4pt}
  \small
  \setlength{\tabcolsep}{5pt}
  \renewcommand{\arraystretch}{1.08}
  \resizebox{\linewidth}{!}{%
  \begin{tabular}{l|c}
  \toprule
  \textbf{Filtering} & \textbf{SearchQA} \\
  \midrule
  RS only      & 54.9 \\
  QF only      & 48.1 \\
  w/o both     & 46.4 \\
  \midrule
  SDAR         & 49.0 \\
  Skill-SD     & 47.8 \\
  \midrule
  \ours{}-9B \textbf{(RS + QF)} & \textbf{58.2} \\
  \bottomrule
  \end{tabular}
  }
\end{minipage}

\end{figure*}

Fig.~\ref{fig:tool} shows a clear gap in long-horizon tool use. 
Qwen3.5-9B-Instruct terminates after only 4.7 rounds on average, often giving premature final answers, whereas \ours{} sustains 18.0 rounds and gathers substantially more evidence for multi-step reasoning. 
Qwen3.5-9B-Instruct also under uses \texttt{visit} for only 0.9 calls, indicating limited evidence access and verification; in contrast, \ours{} performs 5.4 \texttt{visit} calls on average, suggesting stronger document-level grounding. 
We further report an \textit{Advanced Capability Score}, evaluated by an LLM judge over five dimensions: planning, memory maintenance, reasoning, self-correction, and evidence gathering toward the final answer, whose rubric is in Appendix~\ref{app:score}. Qwen3.5-9B-Instruct scores only 19\%, while \ours{} reaches 70\%, demonstrating stronger tool-use strategy.

\subsection{Ablation Studies}

\subsubsection{Impact of State Internalization}

\begin{table}[t]
\centering
\small
\resizebox{0.95\linewidth}{!}{%
\begin{tabular}{l|c}
\toprule
\textbf{Configuration} & \textbf{DeepSearch-Val} \\
\midrule
Qwen3.5-9B-Instruct & 8.5 \\
SFT & 25.0  \\
\midrule
Full pipeline                          & 31.9 \\
\quad w/o state internalization        & 23.5 \\
\quad w/o reflection rewriting         & 16.7 \\
\quad w/o both                         & 14.8 \\
\bottomrule
\end{tabular}
}
\caption{Ablation on trajectory-to-training-data conversion strategies. Results are in percentages averaged over three runs.}
\label{tab:ablation_conversion}
\end{table}

Tab.~\ref{tab:ablation_conversion} studies the conversion of scaffolded trajectories into ReAct-format SFT data. 
Reflection rewriting only masks entity names in environment feedback to prevent answer leakage, while preserving the feedback structure rather than replacing it with oracle answers.

The full pipeline outperforms vanilla SFT on 10K converted trajectories, improving DeepSearch-Val from 25.0 to 31.9, demonstrating the benefit of evolving self-distillation. 
Reflection rewriting is critical: removing it drops performance to 16.7 and 15.7, respectively, as raw reflections contain artifacts such as \texttt{[REFLECTION]} tokens and incomplete environment messages that distort the \texttt{<think>} distribution. 
State internalization provides smaller but consistent gains by injecting planning, memory, and progress tracking into the reasoning trace. 

\subsubsection{Impact of Trajectory Quality Filtering}

Tab.~\ref{tab:ablation_quality_filter} ablates rejection sampling (RS) and trajectory quality filtering (QF) in teacher trajectory generation. 
RS gives the dominant gain, raising SearchQA from 46.4 to 54.9, indicating that answer-correctness verification is the key safeguard for self-distillation.
QF alone brings a smaller gain, but combined with RS in \ours{}-9B, it further improves performance to 58.2 by filtering redundant, weakly aligned, or inconsistent answer-correct traces. 
The clear advantage over OPSD-based~\citep{lu2026self} and Skill-SD~\citep{wang2026skill} suggests that evolving SFT with verified trajectory filtering offers a more stable and effective self-evolving path.

\section{Conclusion}
\label{sec:conclusion}

We introduced \ours{}, a deterministic tool environment built over an offline Wikipedia corpus for deep search agents. 
It provides reproducible observations, entity-level progress verification, and grounded reflection signals, making tool-use trajectories stable and verifiable. 
Building on this environment, \framework{} performs self-distillation by generating verified trajectories and converting scaffold supervision into standard ReAct-format training data. 
This process helps the student agent acquire planning, memory maintenance, self-correction, and evidence-grounded tool use, offering a scalable path for improving tool-use agents from self-generated experience. 
Experiments show that \ours{}-9B substantially improves over its off-the-shelf backbone variant, Qwen3.5-9B, across deep-search and reasoning benchmarks, while remaining competitive with strong open-source agents.

\clearpage
\section*{Limitation}
First, our current environment is built on Wikipedia, which limits coverage and domain diversity. Extending verifiable environments to broader knowledge sources could improve generalization. 
Second, our update rule relies on evolving SFT, while RL-style or OPSD-based updates may further improve generalization and flexibility. However, how to inject higher-level capabilities, such as planning, error recovery, and tool-use strategies, into RL training remains underexplored. We believe addressing these limitations will further broaden the scope of verifiable self-evolution for tool-use agents.

\section*{Ethical Considerations}
This work uses publicly available Wikipedia content, which is licensed under CC-BY-SA, and does not involve private, sensitive, or personally identifiable information. We will release DeepSearch-World and the constructed dataset for research purposes under a CC BY-SA-compatible license, with attribution to Wikipedia contributors.

Our environment is fully offline and deterministic, reducing risks from uncontrolled web access during training. The dataset does not collect private or newly generated personally identifiable information. It may contain publicly available names and biographical facts already present in Wikipedia. However, Wikipedia may contain societal biases, and models trained on it may inherit such biases. Future work should explore more diverse data sources and stronger safety alignment for tool-use agents.

\label{app:ethics}

\bibliography{custom}

\clearpage
\appendix

\section{Details of Data and environment construction.}

We construct a large-scale multi-hop QA dataset of 420K instances designed to require genuine multi-step tool use. The construction pipeline has two phases.

\subsection{Data and Environment Construction}
\label{app:environment_details}

Starting from seed Wikipedia articles, we perform BFS-based exploration of the Wikipedia hyperlink graph:

\begin{enumerate}[leftmargin=*,itemsep=2pt]
    \item \textbf{Knowledge Tree Construction.} For each seed article, we perform breadth-first exploration. At each level, an LLM (Gemini-3-Pro) selects the $b$ most relevant hyperlinks (default $b{=}3$) from the current page. Exploration continues for up to $d_{\max}$ levels (default $d_{\max}{=}4$) or until $p_{\max}$ pages (default $p_{\max}{=}6$) have been visited. Each edge in the tree records the semantic relationship between parent and child articles.

    \item \textbf{Subtree Sampling.} From the complete knowledge tree, we randomly sample 2--4 subtrees. Each subtree defines a set of entities (nodes) and their relationships (edges).

    \item \textbf{Question Generation.} For each sampled subtree, the LLM generates a complex multi-hop question that requires locating and cross-referencing information about all entities in the subtree. The question is designed to be answerable only by consulting multiple Wikipedia articles.

    \item \textbf{Feature Fuzzification.} To prevent memorization and ensure genuine multi-step reasoning, we apply \emph{feature fuzzification}: specific values in the question are replaced with approximate descriptors (e.g., ``2014'' $\to$ ``the 2010s'', ``Beijing, China'' $\to$ ``a major Chinese city''). This forces the agent to search for and verify specific facts rather than pattern-matching from training data.
\end{enumerate}

The tool environment requires a complete, preprocessed Wikipedia corpus:
\begin{enumerate}[leftmargin=*,itemsep=1pt]
    \item Wikipedia XML dumps are processed via WikiExtractor to obtain clean article text.
    \item Extracted articles are merged into a single JSONL file with \texttt{\{id, contents, caption\}} schema.
    \item A SQLite offset index is built for random access by article title.
    \item A Pyserini BM25 index is constructed over the full corpus for keyword retrieval.
    \item Missing text (articles not in the dump) is supplemented via controlled online scraping with rate limiting.
\end{enumerate}
The resulting corpus contains approximately 10 million entries, providing comprehensive coverage of English Wikipedia.

\subsection{Tool Implementation Details}
\label{app:virtual_tool_details}

\paragraph{\texttt{web\_search\_wiki}.}
\texttt{web\_search\_wiki} implements the search interface of the virtual Wikipedia environment. 
It takes a natural-language query as input and returns the top-$k$ retrieved articles, with $k{=}5$ by default. 
The retrieval backend is a Lucene BM25 index constructed with Pyserini~\citep{lin2021pyserini} over the offline Wikipedia corpus. 
Each indexed document contains three fields: \texttt{keyword}, \texttt{caption}, and \texttt{contents}, corresponding to the article title, lead-section summary or feature description, and full article text, respectively. 
For each retrieved article, the tool returns a structured result
\begin{equation}
    r_i = \{\texttt{keyword}_i, \texttt{caption}_i, \texttt{url}_i\},
    \quad i = 1,\ldots,k,
\end{equation}
where \texttt{url} is a deterministic Wikipedia-style URL generated from the normalized article title. 
Thus, the agent receives the same type of information as in a real search engine interface: a ranked list of page titles, snippets, and links, rather than full documents. 
The BM25 index is initialized once and shared across agent workers to reduce loading overhead.

\paragraph{\texttt{visit\_wiki}.}
\texttt{visit\_wiki} implements the page-reading interface. 
It takes a Wikipedia-style URL as input, either returned by \texttt{web\_search\_wiki} or directly generated by the agent, and retrieves the corresponding full article from the offline corpus. 
To support efficient random access, we use a two-level offset-indexed storage design. 
First, a SQLite database maps each normalized article title to its byte offset and length in the JSONL corpus. 
Second, the JSONL corpus stores the full article content. 
Given a URL, the tool extracts and normalizes the article title, looks up its offset and length in SQLite, and then directly seeks to the corresponding byte range in the corpus file, avoiding sequential scanning.

The title normalization pipeline includes underscore-to-space conversion, URL decoding, parenthetical disambiguation handling, and Unicode normalization. 
To reduce I/O overhead, the SQLite index can be loaded into shared memory at startup and reused by parallel agent workers. 
When an article is longer than a predefined threshold, the environment optionally calls a lightweight summarization model to return a condensed page representation containing key facts and evidence; if summarization fails, the tool falls back to deterministic character-level truncation. 
This design preserves the input--output schema of a real browser-visit tool while ensuring deterministic and efficient offline retrieval.

\paragraph{Tool Definition}
In the ReAct framework, each tool is defined through a structured prompt that specifies both its callable format and its semantic capability. This design ensures the language model can reason about tool usage and invoke them appropriately within the \texttt{<tool\_call>...</tool\_call>} block during interaction. Cache is used to prevent search repeatably. In detail, our tools are defined as follows.

\begin{tcolorbox}[title=Tool: \texttt{Web Search}]
\textbf{Description:} Retrieves the top 10 text excerpts from Google's text search engine using one or more search queries.

\textbf{Arguments:}
\begin{itemize}
    \item \texttt{queries} (array of strings): List of search queries. \textit{(Required)}
\end{itemize}
\end{tcolorbox}

\begin{tcolorbox}[title=Tool: \texttt{Visit}]
\textbf{Description:} Visits a given webpage and returns a summary based on a specified goal.

\textbf{Arguments:}
\begin{itemize}
    \item \texttt{url} (string): The target webpage URL. \textit{(Required)} 
    \item \texttt{goal} (string): The goal or information the agent seeks from the webpage. \textit{(Required)}
\end{itemize}
\end{tcolorbox}

\subsection{Teacher Rollout Implementation Details}
\label{app:teacher_rollout_details}

\paragraph{Interaction budget and context management.}
During teacher rollout, the agent is allowed to interact with the virtual environment for up to $T_{\max}$ steps. 
To control context length, we use a sliding-window context strategy: the most recent $w$ interaction steps are retained in full detail, while earlier steps are summarized into the progress state. 
This preserves recent observations and tool-use rationales while preventing long trajectories from exceeding the context window.

\paragraph{Text-based function calling.}
We implement tool use with a text-based function-calling protocol. 
Each tool call is generated as plain text with delimiter tokens specifying the function name, arguments, tool result, and returned observation:
\begin{verbatim}
✿FUNCTION✿: web_search_wiki
✿ARGS✿: {"query": "..."}
✿RESULT✿: [{"keyword": "...", "caption": "..."}]
✿RETURN✿: ...
\end{verbatim}
The delimiters are selected to be unlikely to appear in natural text and are parsed with regular expressions for tool dispatch. 
This protocol is backend-agnostic and allows the same rollout code to be used with models that do not support native function-calling APIs.

\paragraph{Observation truncation.}
Tool observations are truncated when necessary to fit the rollout context. 
For long retrieved pages, the environment either returns a condensed representation or applies deterministic truncation, depending on the configuration described in Appendix~\ref{app:virtual_tool_details}.

\section{Experimental Details}

\subsection{Benchmark Details}
\label{app:benchmark_details}
We evaluate the agent on six search-intensive benchmarks covering English browsing, Chinese browsing, expert-level reasoning, and search-based QA. BrowseComp contains 1,266 English browsing questions designed to test whether agents can persistently locate hard-to-find and entangled information on the web. BrowseComp-ZH extends this setting to the Chinese web with 289 natively constructed multi-hop questions across 11 domains, requiring agents to handle Chinese-specific retrieval challenges such as fragmented platforms, implicit expressions, and heterogeneous information sources. HLE evaluates frontier-level academic reasoning with 2,158 English multiple-choice questions spanning humanities, science, and mathematics. GAIA-Text contains 103 text-only questions from the GAIA development split, focusing on real-world multi-step queries that require web search and reasoning. xBench-DeepSearch contains 100 Chinese deep-search questions for evaluating web search and reasoning ability, with encrypted public data to mitigate benchmark contamination. Search-QA follows prior search-based QA evaluations and includes both single-hop retrieval datasets, such as NQ, TriviaQA, and PopQA, and multi-hop QA datasets, such as HotpotQA, 2Wiki, MuSiQue, and Bamboogle, testing whether agents can retrieve and compose evidence across one or multiple supporting documents.

\subsection{Baseline Details}
\label{app:baseline_details}
We compare against proprietary models, open-source backbones, and open-source search agents under two evaluation settings.
In the \emph{Direct Reasoning} setting, models answer in a single pass without external retrieval or tool use.
This group includes GPT-5, Claude-4-Sonnet, Claude-3.7-Sonnet, OpenAI Deep Research, OpenAI-o3, and the Qwen3.5-9B-Instruct backbone.

In the \emph{Open Source Agents} setting, we compare with recent search and deep-research agents at comparable model scales, including R1-Searcher-7B, Search-R1-7B, ZeroSearch-7B, ASearcher-7B, DeepResearcher-7B, PokeeResearch-7B, WebSailor-7B, WebExplorer-8B, Marco-DR-8B, MiroThinker-v1.0-8B, and DeepDive-9B.
Since publicly reported deep-search agents built exactly on 9B backbones are still limited, we include nearby 7B--8B agents to provide a broader and more informative comparison within the same small-model regime.
This comparison is therefore not intended as a strict parameter-matched contest, but as a practical assessment against recent open-source agents with similar deployment budgets.
Results for R1-Searcher-7B, Search-R1-7B, ZeroSearch-7B, ASearcher-7B, and DeepResearcher-7B are taken from the PokeeResearch 7B-scale re-evaluation.

\subsection{Training Hyperparameter}
\label{app:train_details}

We run the self-evolving loop for $R{=}15$ rounds with the following configuration:
\begin{itemize}[leftmargin=*,itemsep=1pt]
    \item \textbf{Batch size}: $B{=}10{,}000$ questions per generation batch.
    \item \textbf{Correct threshold}: $C_{\min}{=}4{,}000$ correct trajectories to trigger training.
    \item \textbf{IS target size}: $N_{\text{target}}{=}4{,}000$ trajectories after importance-sampling mixing.
    \item \textbf{Decay factor}: $\gamma{=}0.5$ for exponential decay across rounds.
    \item \textbf{Training}: Llama Factory~\citep{zheng2024llamafactory} with 1 epoch per round, learning rate $5 \times 10^{-6}$, cosine schedule, 10\% warmup, DeepSpeed ZeRO-2, BFloat16 precision, max sequence length 32,768.
    \item \textbf{Template}: \texttt{qwen3\_5\_nothink} (strips \texttt{<think>} tokens during training to focus on trajectory structure).
    \item \textbf{Agent}: $T_{\max}{=}30$ steps, recent window $w{=}3$, observation truncation at 8,192 characters.
\end{itemize}

\subsection{Asynchronous Training Efficiency}

The asynchronous design enables significant throughput gains. On a single 8-GPU H20 node with Qwen3.5-9B, the system achieves:
\begin{itemize}[leftmargin=*,itemsep=1pt]
    \item \textbf{Generation throughput}: Processing ${\sim}$10,000 questions per generation batch with vLLM (TP=4 on GPUs 0--3, or TP=8 when summary is remote).
    \item \textbf{Training throughput}: Full SFT on 4,000 trajectories completes in one epoch during the time the next generation batch is running.
    \item \textbf{End-to-end}: A complete 5-round self-evolving loop completes in approximately 2--3 days on a single 2-node H20 cluster.
\end{itemize}
The critical design choice is that generation \emph{never blocks on training}: the orchestrator continues producing trajectories while training runs in the background, and model swaps occur only at round boundaries.

\subsection{Advanced Capability Score.}
\label{app:score}
We use an LLM judge to assess each trajectory along five dimensions: 
(1) planning, whether the agent decomposes the question into actionable subgoals; 
(2) memory, whether it maintains and updates useful intermediate evidence; 
(3) reasoning, whether it connects evidence across hops coherently; 
(4) self-correction, whether it revises failed searches or wrong assumptions; and 
(5) evidence gathering, whether it retrieves sufficient supporting information before answering. 
The final score is the average pass rate across these five dimensions.

\section{Prompts}
\subsection{Trajectory Generation}

We use this prompt when obtaining the trajectory with correct responses using reject sampling.

\begin{tcolorbox}[breakable,title=\texttt{Prompt: Evaluation of Reject Sampling}]
\textbf{Task:} Please evaluate whether the model’s answer is correct based on the given question, standard answer, and model-predicted answer. Rate the result as:
\begin{itemize}
  \item A: [Correct]
  \item B: [Incorrect]
  \item C: [Not Attempted]
\end{itemize}
Return only the letter “A”, “B”, or “C”, with no additional text.

\bigskip

\textbf{Examples of [Correct] responses:}
\begin{verbatim}
Question: What are the names of Barack Obama's children?
Standard Answer: Malia Obama and Sasha Obama
Model Prediction 1: Malia Obama and Sasha Obama
Model Prediction 2: Malia and Sasha
Model Prediction 3: Most people would say it's Malia and Sasha, 
but I'm not sure.
Model Prediction 4: Barack Obama has two daughters, named 
Malia Ann and Natasha Marian, but usually referred to as Malia 
Obama and Sasha Obama. Malia was born on July 4, 1998, and 
Sasha was born on June 10, 2001.
\end{verbatim}

\medskip

\textbf{Examples of [Incorrect] responses:}
\begin{verbatim}
Question: What are Barack Obama's children's names?
Standard Answer: Malia Obama and Sasha Obama
Model Prediction 1: Malia
Model Prediction 2: Malia, Sasha, and Susan
Model Prediction 3: Barack Obama has no children
Model Prediction 4: I think it's Malia and Jackie.
Model Prediction 5: Although I don't know their exact names, I 
can say Barack Obama has three children. 
Model Prediction 6: You might refer to Betsy and Olivia...
\end{verbatim}

\medskip

\textbf{Examples of [Not Attempted] responses:}
\begin{verbatim}
Question: What are Barack Obama's children's names?
Standard Answer: Malia Obama and Sasha Obama
Model Prediction 1: I don't know.
Model Prediction 2: I need more context about which Obama you 
refer to.
Model Prediction 3: Without checking online, I can't answer 
this question.
Model Prediction 4: Barack Obama has two children. I know one 
is named Malia, but I'm not sure of the other's name.
\end{verbatim}

\bigskip

\textbf{Notes:}
\begin{itemize}
  \item Numerical answers: near matches (e.g.\ “3518” vs. “3518.17”) are [Correct]; wrong numbers are [Incorrect]; vague ranges are [Not Attempted].
  \item If the standard answer has extra detail, the prediction only needs the part asked by the question.
  \item If missing details can be inferred from the question, treat as [Correct].
\end{itemize}

\bigskip

\textbf{Now evaluate:}
\begin{verbatim}
Question: {question}
Standard Answer: {target}
Predicted Answer: {predicted_answer}
\end{verbatim}

\textbf{Return:} A, B, or C
\end{tcolorbox}

\subsection{Quality Filter}
\label{app:prompt_filter}
This prompt is used to guide model in verifying whether a given trajectory is logically sound and consistent with the task requirements.

\begin{tcolorbox}[breakable,title=\texttt{Prompt: Tool Call Rationality Evaluation}]
\textbf{Role:} You are a professional AI interaction quality assessor. Your core task is to analyze dialogue snippets between a user and an AI assistant that include a \texttt{<tool\_call>} tag followed by a \texttt{<think>} tag.

\bigskip

\textbf{Task:} Judge whether the tool call (\texttt{<tool\_call>}) is \emph{reasonable} according to the three criteria defined below. “Reasonable” means the call is necessary, directly driven by the user’s query, efficient, precise, non-redundant, and conforms to specifications. Also evaluate the thought process (\texttt{<think>}) for logical accuracy and to ensure no guessing or fabrication.

\bigskip

\textbf{Evaluation Criteria:}
\begin{enumerate}
  \item \textbf{Information Non-Redundancy:}  
    The requested information or action in the tool call is \emph{not} already provided or easily derivable from prior dialogue, the user’s current question, or the assistant’s previous answers.  
    \emph{Check:} Is there any overlap or repeated request?
  \item \textbf{Goal Alignment:}  
    The tool call’s purpose and expected result directly serve the user’s explicit intent or core need in this turn.  
    \emph{Check:} Does it advance the user’s main objective?
  \item \textbf{Logical Reasoning and Accuracy:}  
    The assistant’s thought process shows clear, correct logic and reliable grounding -- no unfounded guesses or fabrications. The \texttt{<think>} section should be concise.  
    \emph{Check:} Is the reasoning well-structured and evidence-based?
\end{enumerate}

\bigskip

\textbf{Instruction:}  
Compare the user’s question and the model’s generated snippet (including \texttt{<tool\_call>} and \texttt{<think>}). If \emph{all} criteria are met, output:
\begin{verbatim}
A
\end{verbatim}
Otherwise (any criterion unmet or room for improvement), output:
\begin{verbatim}
B
\end{verbatim}

\bigskip

\textbf{User Question:}  
\{query\}

\medskip

\textbf{Model Generation:}  
\{model\_gen\}
\end{tcolorbox}

\subsection{ReAct Trajectories}

\begin{tcolorbox}[breakable,title=Case Trajectory]
<think> thinking process here </think>\\
<tool\_call>\\
{"name": "tool name here", "arguments": {"parameter name here": parameter value here, "another parameter name here": another parameter value here, ...}}\\
</tool\_call>\\
<tool\_response>\\
tool\_response here\\
</tool\_response>\\
(more thinking processes, tool calls and tool responses here)\\
<think> thinking process here </think>\\
<answer> answer here </answer>
\end{tcolorbox}

\subsection{Planning User Prompt}

We use this prompt to update the progress state during trajectory generation of teacher scaffold.

\begin{tcolorbox}[breakable,title=\texttt{Prompt: State Update}]

\bigskip

\textbf{\#\# Target Task}
\begin{verbatim}
{task}
\end{verbatim}

\textbf{\#\# Recent Steps}
\begin{verbatim}
{recent_steps_str}
\end{verbatim}

\textbf{\#\# Current Progress State}
\begin{verbatim}
{state}
\end{verbatim}

\textbf{\#\# Output}

Based on the observations above, output the updated Progress State JSON.

Move completed items to \texttt{completed\_list}, adjust \texttt{todo\_list} to reflect what still needs to be done, record new facts in \texttt{information}, add lessons to \texttt{experience}.

Output ONLY the JSON block, no explanation.

\begin{verbatim}
```json
{"completed_list": [...], "todo_list": [...], "experience": [...], "information": [...]}

\end{verbatim}
\end{tcolorbox}

\subsection{Action User Prompt}

We use this prompt to call tools during trajectory generation of teacher scaffold.

\begin{tcolorbox}[breakable,title=\texttt{Prompt: Action}]
\begin{verbatim}
Please generate your response using EXACTLY ONE of the two formats below:

**Format A — call a tool (when you still need information):**
<think> I know: [what is confirmed from previous steps]. I still need: [what specific information is missing]. Next: [which tool and why]. </think>
<tool_call>
{"name": "tool_name_here", "arguments": {"param1": "value1", "param2": "value2"}}
</tool_call>

**Format B — give the final answer (ONLY when the confirmed answer is already in hand from tool results):**
<think> I know: [the specific fact/number/name from page content that answers the task]. No more searches required. </think>
<answer>DIRECT ANSWER HERE</answer>

CRITICAL: Every response MUST start with <think>. A response without <think> is INVALID.
IMPORTANT: Do NOT use Format B to describe next steps or search plans. If you still need to search, use Format A.
IMPORTANT: Do NOT output `[REFLECTION]` in your response. Reflection messages are injected by the environment automatically — never generate them yourself.
IMPORTANT: If your last action was `web_search_wiki`, your ONLY valid next action is `visit_wiki`. You MUST call `visit_wiki` now — do NOT answer, do NOT search again.
\end{verbatim}
\end{tcolorbox}

\subsection{Ending User Prompt}

We use this prompt to generate the final answer from the completed trajectory of teacher scaffold.

\begin{tcolorbox}[breakable,title=\texttt{Prompt: Final Answer Generation}]
\begin{verbatim}
## Available Information
- `Target Task`: The specific task to be completed.
- `Recent Steps`: The most recent actions taken by the agent.
- `Previous Progress State`: A JSON representation of the task's progress, including key information and milestones.
- `Final Step`: The last action taken and its observation.

## Guidelines
1. **Final Result**: Carefully examine the outputs from the previous steps to decide the final output.
2. **Output Rules**: Your final output should be a number OR as few words as possible OR a comma separated list of numbers and/or strings. Do NOT include any unnecessary information in the output.
    - **Number**: If you are asked for a number, directly output the number itself. Don't use comma to write your number. Be careful about what the question is asking, for example, the query might ask "how many thousands", in this case, you should properly convert the number if needed. Nevertheless, do NOT include the units (like $, %, km, thousands and so on) unless specified otherwise.
    - **String**: If you are asked for a string, don't use articles, neither abbreviations (e.g. for cities), and write the digits in plain text unless specified otherwise.
    - **List**: If you are asked for a comma separated list, apply the above rules depending of whether the element to be put in the list is a number or a string.
Please generate your response, your reply should strictly follow the format:
<think> Scan the observations and information fields for the specific fact, number, or name that directly answers the Target Task. State what you found and verify it matches the required answer format. </think>
<answer>DIRECT ANSWER HERE</answer>
\end{verbatim}
\end{tcolorbox}

\subsection{System Prompt}

\begin{tcolorbox}[breakable,title=\texttt{System Prompt of teacher}]
\begin{verbatim}
"""You are the action module of the Agent. Your role is to select and call the appropriate tool for the next step.

## Available Information
- `Target Task`: The specific task to be completed.
- `Recent Steps`: The most recent actions taken by the agent.
- `Previous Progress State`: A JSON representation of the task's progress, including key information and milestones.
- `Tool Definitions`
- `Tool outputs`

## Agent Loop — message sequence per turn
Each tool-call turn produces the following messages in the conversation history:

1. `assistant`: Your response — `<think>...</think>` then `<tool_call>...</tool_call>`.
2. `user`: The tool result — `<tool_response>...</tool_response>`.
3. `user`: A `[REFLECTION]` message — environment-injected progress feedback.
4. `user`: A `[STATE_UPDATE]` prompt — asking you (in your state-tracker role) to update the Progress State.
5. `assistant`: Your state-update response — a JSON block with the updated Progress State.

**When you are acting as the action module (this prompt)**, you produce messages 1 and 5.
Messages 2 and 3 are injected by the environment. Message 4 is a separate prompt sent to you in a state-tracker role.
Do NOT generate `[REFLECTION]` or `[STATE_UPDATE]` in your action response.

## Rules
1. Call one tool at a time; wait for its result before deciding the next step.
2. If a tool call fails or returns irrelevant results, try a different query or tool — do not repeat the same call.
3. Each tool call is stateless — its internal state is discarded after return. Pass all necessary context explicitly in the arguments.
4. Use `<answer>` ONLY when you already have the confirmed, specific answer (a fact, name, number, etc.) obtained from tool results. NEVER put next steps, search plans, or action descriptions inside `<answer>`.
5. `[REFLECTION]` messages are **system-injected** progress-tracking signals — they are added automatically by the environment after tool calls. **You must NEVER generate or output `[REFLECTION]` yourself.** Keywords in "Found target keyword X" or "remaining targets" are intermediate research targets to look up — they are NOT the answer. Only output `<answer>` after extracting the actual fact from page content.
6. **`<think>` is MANDATORY**: You MUST output `<think>...</think>` BEFORE every `<tool_call>` or `<answer>`. A response without `<think>` is invalid and will be rejected.

## Tool Call Guidelines
- **Mandatory two-step pattern — NO EXCEPTIONS**: `web_search_wiki` → `visit_wiki` → extract fact → answer.
  - After `web_search_wiki` returns, you MUST call `visit_wiki` next. Always. No exceptions.
  - After `visit_wiki` returns, read the content and extract the fact. Only then may you answer.
  - NEVER answer from a caption. NEVER call `web_search_wiki` twice in a row.
- Only facts confirmed from a visited page may be used to answer. Never answer from captions alone.

## Complete Example

Below is a full trajectory. Each message is labeled with its role. This is exactly the sequence that appears in the conversation history.

### Turn 1 — Search

[assistant] ← YOUR output (action module)
<think> I know: nothing yet. I still need: which club Graham Page played for, and its link to Sydney Swans. Next: web_search_wiki for Graham Page (footballer). </think>
<tool_call>
{"name": "web_search_wiki", "arguments": {"query": "Graham Page (footballer)"}}
</tool_call>

[user] ← tool result (environment-injected)
<tool_response>[{"keyword": "Graham Page (footballer)", "score": 15.28, "caption": "Graham Page played Australian rules football for South Melbourne in the Victorian Football League.", "urls": ["https://en.wikipedia.org/wiki/Graham_Page_(footballer)"]}]</tool_response>

[user] ← reflection (environment-injected)
[REFLECTION] web_search_wiki: Found target keyword "Graham Page", progress updated.
Current progress: 1/2.

[user] ← state update prompt (environment-injected, sent to you in state-tracker role)
[STATE_UPDATE]
Task: Which club did Graham Page play for before it became the Sydney Swans?
State: {"completed_list": [], "todo_list": ["Search: Graham Page (footballer)", "Search: Sydney Swans", "Confirm which club became Sydney Swans"], "experience": [], "information": []}
Output the updated Progress State JSON.

[assistant] ← YOUR output (state-tracker role)
```json
{"completed_list": ["Searched Graham Page (footballer) — page found"], "todo_list": ["Visit Graham Page Wikipedia page to confirm club name", "Search: Sydney Swans to confirm renaming history"], "experience": [], "information": ["Graham Page played for South Melbourne (caption, unconfirmed)"]}
Turn 2 — Visit

[assistant] ← YOUR output (action module)
<think> I know: caption says South Melbourne, but captions are too brief to confirm. I still need: full page content to confirm the club name and its link to Sydney Swans. Next: visit_wiki on the Graham Page URL. </think>
<tool_call>
{"name": "visit_wiki", "arguments": {"url": "https://en.wikipedia.org/wiki/Graham_Page_(footballer)", "goal": "Find which club Graham Page played for and its connection to Sydney Swans"}}
</tool_call>

[user] ← tool result (environment-injected)
<tool_response>Graham Page (born 1 June 1939) is a former Australian rules footballer who played for South Melbourne in the Victorian Football League (VFL). South Melbourne relocated to Sydney in 1982 and was renamed the Sydney Swans.</tool_response>

[user] ← reflection (environment-injected)
[REFLECTION] visit_wiki: Found target keyword "Sydney Swans", progress updated.
Current progress: 2/2.

[user] ← state update prompt (environment-injected, sent to you in state-tracker role)
[STATE_UPDATE]
Task: Which club did Graham Page play for before it became the Sydney Swans?
State: {"completed_list": ["Searched Graham Page (footballer) — page found"], "todo_list": ["Visit Graham Page Wikipedia page to confirm club name", "Search: Sydney Swans to confirm renaming history"], "experience": [], "information": ["Graham Page played for South Melbourne (caption, unconfirmed)"]}
Output the updated Progress State JSON.

[assistant] ← YOUR output (state-tracker role)

{"completed_list": ["Confirmed: Graham Page played for South Melbourne", "Confirmed: South Melbourne became Sydney Swans in 1982"], "todo_list": [], "experience": [], "information": ["Graham Page played for South Melbourne (VFL)", "South Melbourne → Sydney Swans in 1982"]}
After many turns
Turn x — Answer (all facts confirmed from page content)

[assistant] ← YOUR output (action module)
<think> I know: the page confirms Graham Page played for South Melbourne, which became the Sydney Swans in 1982. I have the specific fact needed to answer the question. No more searches required. </think>
<answer>South Melbourne</answer>
"""
\end{verbatim}
\end{tcolorbox}

\begin{tcolorbox}[breakable,title=\texttt{System Prompt of student}]
\begin{verbatim}
You are a multi-step QA agent that plans research, calls tools, and produces concise final answers.

## Rules
1. Call one tool at a time; wait for its result before deciding the next step.
2. If a tool call fails or returns irrelevant results, try a different query or tool — do not repeat the same call.
3. Each tool call is stateless — pass all necessary context explicitly in the arguments.
4. Use `<answer>` ONLY when you already have the confirmed answer from tool results. NEVER put next steps or search plans inside `<answer>`.

## Progress State
Four fields (ALL writable — update every round):
- completed_list: Finished steps and confirmed findings.
- todo_list: **Living checklist** — remove finished items, add new follow-ups, rewrite failing searches, reorder by urgency. Never leave it unchanged after a tool result; never leave it empty while unsolved.
- experience: Lessons learned or failed strategies.
- information: Confirmed facts extracted from visited pages only.

## Tool Call Guidelines
- **Mandatory pattern**: `web_search_wiki` (captions only — never sufficient) → `visit_wiki` (full page) → extract fact → answer. Never skip `visit_wiki`.
- Only facts from a visited page may be used to answer. Never answer from captions alone.

## Final Answer Format
- **Number**: digits only, no commas, no units unless specified.
- **String**: no articles, no abbreviations, digits in plain text unless specified.
- **List**: comma-separated, applying the above rules per element.
- `<answer>` must contain the direct fact/number/name — not process descriptions.\
\end{verbatim}
\end{tcolorbox}

\section{Case Study}
\label{app:case}

\begin{tcolorbox}[breakable,title= ReAct-style Trajectory in \ours{} of validation set]
\textcolor{black}{\textbf{Question:} A now-defunct performing organization was founded in the first decade of the twentieth century after the surplus from a visiting company's successful local season was used to establish it. Several years later, it moved beyond its original supporting role and mounted its first independent production, choosing a work whose nineteenth-century Paris premiere took place at a theatre later associated with the city's mid-century cultural redevelopment. In the interwar period, under a composer-organist who had recently created a society for promoting contemporary music, the organization introduced to Britain an opera by an eighteenth-century composer. Decades later, after its annual productions moved to a theatre housed within a national conservatoire, the organization staged a compact early-twentieth-century opera that had not been produced in the United Kingdom for about eight decades. What is the name of that opera and its composer?}\\

\colorbox{gray!70}{%
    \parbox{\dimexpr\textwidth-2\fboxsep}{%
    \centering
        \textbf{Step 1}%
    }%
}
\textcolor{purple1}{\textbf{<think>} \\
\textbf{Progress state.}\\
\texttt{\{}\\
\texttt{"completed\_list": [],}\\
\texttt{"todo\_list": ["Identify the defunct performing organization", "Verify its first independent production", "Identify the interwar conductor and British premiere", "Find the later venue move", "Recover the revived early-twentieth-century opera and composer"],}\\
\texttt{"experience": [],}\\
\texttt{"information": []}\\
\texttt{\}}\\
\textbf{Reflection.}\\
No prior evidence is available. The first step should target the founding clue because it combines a rare set of constraints: early-twentieth-century founding, surplus from a visiting company's profitable local season, and an originally supporting role.\\
\textbf{Action.}\\
Search for the organization using the founding mechanism rather than the final opera.\\
\textbf{</think>}}\\
\textcolor{black}{\textbf{<tool\_call>} \\
\{"query": "defunct performing organization founded surplus profits visiting opera company successful local season original supporting role"\}
\\
\textbf{</tool\_call>}}\\
\textcolor{black}{
\textbf{<tool\_response>}\\
Search results identify the Glasgow Grand Opera Society. It was founded in 1906 after profits from the Moody-Manners Opera Company's successful Glasgow season were used to create the society. Its original purpose was to augment touring grand opera companies' choruses.\\
\textbf{</tool\_response>}
}\\

\colorbox{gray!70}{%
    \parbox{\dimexpr\textwidth-2\fboxsep}{%
    \centering
        \textbf{Step 2}%
    }%
}
\textcolor{purple1}{\textbf{<think>} \\
\textbf{Progress state.}\\
\texttt{\{}\\
\texttt{"completed\_list": ["Identified candidate organization: Glasgow Grand Opera Society"],}\\
\texttt{"todo\_list": ["Verify its first independent production", "Identify the interwar conductor and British premiere", "Find the later venue move", "Recover the revived early-twentieth-century opera and composer"],}\\
\texttt{"experience": ["Founding clues are strong enough to locate the organization, but the candidate must be validated through later production-history clues."],}\\
\texttt{"information": ["The organization was founded in 1906 using profits from the Moody-Manners Opera Company's successful Glasgow season.", "Its original role was to augment touring opera choruses."]}\\
\texttt{\}}\\
\textbf{Reflection.}\\
The first search resolves the likely institution, but the answer depends on a later revived work. Before moving forward, I should verify the clue about its first independent production to avoid locking onto a false institutional match.\\
\textbf{Action.}\\
Search the candidate organization together with the first independent production clue.\\
\textbf{</think>}}\\
\textcolor{black}{\textbf{<tool\_call>} \\
\{"query": "Glasgow Grand Opera Society first independent production 1911 Faust"\}
\\
\textbf{</tool\_call>}}\\
\textcolor{black}{\textbf{<tool\_response>}\\
The society mounted its first independent production in 1911 with Gounod's \emph{Faust}.\\
\textbf{</tool\_response>}}\\

\colorbox{gray!70}{%
    \parbox{\dimexpr\textwidth-2\fboxsep}{%
    \centering
        \textbf{Step 3}%
    }%
}
\textcolor{purple1}{\textbf{<think>} \\
\textbf{Progress state.}\\
\texttt{\{}\\
\texttt{"completed\_list": ["Identified candidate organization: Glasgow Grand Opera Society", "Verified first independent production: Gounod's Faust"],}\\
\texttt{"todo\_list": ["Verify the Paris-premiere clue for Faust", "Identify the interwar conductor and British premiere", "Find the later venue move", "Recover the revived early-twentieth-century opera and composer"],}\\
\texttt{"experience": ["Institutional clues should be cross-checked through production records before using them as anchors."],}\\
\texttt{"information": ["Glasgow Grand Opera Society mounted Gounod's Faust in 1911 as its first independent production."]}\\
\texttt{\}}\\
\textbf{Reflection.}\\
The first-production clue matches, but the question also encodes the work through its Paris premiere. I should verify that \emph{Faust} fits this clue, strengthening the chain before proceeding.\\
\textbf{Action.}\\
Search for the original premiere venue of Gounod's \emph{Faust}.\\
\textbf{</think>}}\\
\textcolor{black}{\textbf{<tool\_call>} \\
\{"query": "Gounod Faust original version premiered Théâtre Lyrique Paris 1859"\}
\\
\textbf{</tool\_call>}}\\
\textcolor{black}{\textbf{<tool\_response>}\\
Gounod's \emph{Faust} premiered in Paris in 1859 at the Théâtre Lyrique.\\
\textbf{</tool\_response>}}\\

\colorbox{gray!70}{%
    \parbox{\dimexpr\textwidth-2\fboxsep}{%
    \centering
        \textbf{Step 4}%
    }%
}
\textcolor{purple1}{\textbf{<think>} \\
\textbf{Progress state.}\\
\texttt{\{}\\
\texttt{"completed\_list": ["Identified candidate organization: Glasgow Grand Opera Society", "Verified first independent production: Gounod's Faust", "Verified Paris-premiere clue for Faust"],}\\
\texttt{"todo\_list": ["Identify the interwar conductor and British premiere", "Find the later venue move", "Recover the revived early-twentieth-century opera and composer"],}\\
\texttt{"experience": ["The organization is now strongly grounded by both founding and production-history evidence."],}\\
\texttt{"information": ["Gounod's Faust premiered at the Théâtre Lyrique in Paris in 1859."]}\\
\texttt{\}}\\
\textbf{Reflection.}\\
The candidate organization remains consistent. The next unresolved clue is the interwar conductor: a composer-organist connected to a contemporary-music society and a British premiere of an eighteenth-century composer's opera.\\
\textbf{Action.}\\
Search for the organization's interwar conductor and the associated British premiere.\\
\textbf{</think>}}\\
\textcolor{black}{\textbf{<tool\_call>}\\
\{"query": "Glasgow Grand Opera Society composer organist contemporary music society British premiere Mozart opera"\}\\
\textbf{</tool\_call>}}\\
\textcolor{black}{\textbf{<tool\_response>}\\
Erik Chisholm, a composer and organist, founded the Active Society for the Propagation of Contemporary Music and became musical director and conductor of Glasgow Grand Opera Society in the 1930s. In 1934, the society gave the British premiere of Mozart's \emph{Idomeneo}.\\
\textbf{</tool\_response>}}\\

\colorbox{gray!70}{%
    \parbox{\dimexpr\textwidth-2\fboxsep}{%
    \centering
        \textbf{Step 5}%
    }%
}
\textcolor{purple1}{\textbf{<think>} \\
\textbf{Progress state.}\\
\texttt{\{}\\
\texttt{"completed\_list": ["Identified candidate organization: Glasgow Grand Opera Society", "Verified first independent production: Gounod's Faust", "Verified Paris-premiere clue for Faust", "Identified interwar conductor: Erik Chisholm", "Verified British premiere: Mozart's Idomeneo"],}\\
\texttt{"todo\_list": ["Find the later venue move", "Recover the revived early-twentieth-century opera and composer"],}\\
\texttt{"experience": ["When a clue describes a person indirectly, linking the person to both institutional role and event outcome prevents accidental entity drift."],}\\
\texttt{"information": ["Erik Chisholm founded the Active Society for the Propagation of Contemporary Music.", "He conducted Glasgow Grand Opera Society in the 1930s.", "The society gave the British premiere of Mozart's Idomeneo in 1934."]}\\
\texttt{\}}\\
\textbf{Reflection.}\\
The historical chain is complete up to the interwar clue. The remaining clue asks about a later production after annual performances moved to a conservatoire theatre. I should search for the society's later venue and production history.\\
\textbf{Action.}\\
Search for the venue move and the early-twentieth-century opera revived after the move.\\
\textbf{</think>}}\\
\textcolor{black}{\textbf{<tool\_call>} \\
\{"query": "Glasgow Grand Opera Society annual productions moved New Athenaeum Theatre 1994 1905 opera first UK production 80 years"\}
\\
\textbf{</tool\_call>}}\\
\textcolor{black}{\textbf{<tool\_response>}\\
From 1993, the society's annual productions moved to the New Athenaeum Theatre at the Royal Conservatoire of Scotland. In 1994, it performed Franco Leoni's 1905 opera \emph{L'Oracolo}, described as the first production of that opera in the United Kingdom for 80 years.\\
\textbf{</tool\_response>}}\\

\colorbox{gray!70}{%
    \parbox{\dimexpr\textwidth-2\fboxsep}{%
    \centering
        \textbf{Step 6}%
    }%
}
\textcolor{purple1}{\textbf{<think>} \\
\textbf{Progress state.}\\
\texttt{\{}\\
\texttt{"completed\_list": ["Identified candidate organization: Glasgow Grand Opera Society", "Verified first independent production: Gounod's Faust", "Verified Paris-premiere clue for Faust", "Identified interwar conductor: Erik Chisholm", "Verified British premiere: Mozart's Idomeneo", "Found later venue move: New Athenaeum Theatre", "Found candidate final opera: L'Oracolo by Franco Leoni"],}\\
\texttt{"todo\_list": ["Independently verify the final opera title and composer"],}\\
\texttt{"experience": ["The final clue is likely solved, but final answers should be independently verified when both title and composer are requested."],}\\
\texttt{"information": ["Annual productions moved to the New Athenaeum Theatre in 1993.", "In 1994, the society performed Franco Leoni's 1905 opera L'Oracolo.", "The production was described as the first UK production of the opera for 80 years."]}\\
\texttt{\}}\\
\textbf{Reflection.}\\
The final answer appears in the society's production history, but I should verify \emph{L'Oracolo}'s composer and date from an opera-specific source.\\
\textbf{Action.}\\
Search directly for \emph{L'Oracolo} and Franco Leoni.\\
\textbf{</think>}}\\
\textcolor{black}{\textbf{<tool\_call>} \\
\{"query": "L'Oracolo opera Franco Leoni 1905 composer"\}
\\
\textbf{</tool\_call>}\\
\textcolor{black}{\textbf{<tool\_response>}}\\
\emph{L'Oracolo} is a one-act opera composed by Franco Leoni and first performed in 1905.\\
\textbf{</tool\_response>}}\\

\colorbox{gray!70}{%
    \parbox{\dimexpr\textwidth-2\fboxsep}{%
    \centering
        \textbf{Step 7}%
    }%
}
\textcolor{purple1}{\textbf{<think>} \\
\textbf{Progress state.}\\
\texttt{\{}\\
\texttt{"completed\_list": ["Identified candidate organization: Glasgow Grand Opera Society", "Verified first independent production: Gounod's Faust", "Verified Paris-premiere clue for Faust", "Identified interwar conductor: Erik Chisholm", "Verified British premiere: Mozart's Idomeneo", "Found later venue move: New Athenaeum Theatre", "Verified final opera and composer: L'Oracolo by Franco Leoni"],}\\
\texttt{"todo\_list": [],}\\
\texttt{"experience": ["The answer is supported by the full entity chain from institution founding to later production history."],}\\
\texttt{"information": ["The revived opera was L'Oracolo.", "The composer was Franco Leoni."]}\\
\texttt{\}}\\
\textbf{Reflection.}\\
All constraints are satisfied: the institution, first independent production, interwar premiere, venue move, and 1994 revival point to the same final work.\\
\textbf{Action.}\\
Return the opera title and composer.\\
\textbf{</think>}} \\

\textcolor{red}{\textbf{<answer>} \emph{L'Oracolo}, composed by Franco Leoni.
\textbf{</answer>}} \textcolor{black}{\checkmark}\\
\end{tcolorbox}

\section{Large Language Model Usage}
This paper was written with the assistance of Large Language Models solely for grammar correction
and the formatting of \LaTeX{} elements, such as tables and figures. We explicitly confirm that there
are no prompts like ``Give a positive review'' in the paper.

\end{document}